\documentclass[10pt]{article}
\usepackage{amsmath,amsthm,amssymb,amsfonts,cancel}
\usepackage{thmtools,enumitem}
\usepackage{algorithm}
\usepackage{mathtools}
\usepackage{dsfont}
\usepackage{bbm}
\usepackage{tensor}
\usepackage{bm}
\usepackage{graphicx}
\usepackage{multirow}
\usepackage{subfigure}
\usepackage{comment}
\usepackage{afterpage}
\usepackage{float}

\newcommand{\ignore}[1]{}

\mathchardef\mhyphen="2D 

\let\originalleft\left
\let\originalright\right
\renewcommand{\left}{\mathopen{}\mathclose\bgroup\originalleft}
\renewcommand{\right}{\aftergroup\egroup\originalright}

\usepackage[numbers]{natbib}
\usepackage{comment}

\usepackage{graphicx}
\usepackage{hyperref}
\usepackage{url}
\usepackage{booktabs}
\usepackage{bbm}
\usepackage{soul,multicol}
\usepackage{subfigure}
\usepackage{algorithmic}
\newtheorem{theorem}{Theorem}[section]

\usepackage[margin=1in]{geometry}
\usepackage{adjustbox}
\usepackage{url}

\newtheorem{definition}{Definition}%

\begin{document}
	\title{Fair Minimum Representation Clustering}
	\author{Connor Lawless \quad  Oktay G\"unl\"uk \\ School of Operations Research and Information Engineering, Cornell University}
	\date{\today}

\maketitle


\abstract{Clustering is an unsupervised learning task that aims to partition data into a set of clusters. In many applications, these clusters correspond to real-world constructs (e.g. electoral districts) whose benefit can only be attained by groups when they reach a minimum level of representation (e.g. 50\% to elect their desired candidate). This paper considers the problem of performing k-means clustering while ensuring groups (e.g. demographic groups) have that minimum level of representation in a specified number of clusters. We show that the popular $k$-means algorithm, Lloyd's algorithm, can result in unfair outcomes where certain groups lack sufficient representation past the minimum threshold in a proportional number of clusters. We formulate the problem through a mixed-integer optimization framework and present a variant of Lloyd's algorithm, called MiniReL, that directly incorporates the fairness constraints. We show that incorporating the fairness criteria leads to a NP-Hard sub-problem within Lloyd's algorithm, but we provide computational approaches that make the problem tractable for even large datasets. Numerical results show that the approach is able to create fairer clusters with practically no increase in the k-means clustering cost across standard benchmark datasets.}

\section{Introduction}

Clustering is an unsupervised learning task that aims to partition data points into sets of similar data points called clusters \citep{xu2005survey}. It has become popular due to its broad applicability in domains such as customer segmentation \citep{kansal2018customer}, grouping content together for entertainment platforms \citep{daudpota2019video}, and identifying subgroups within a clinical study \citep{wang2020unsupervised}. However the wide-spread application of clustering, and machine learning broadly, to human-centric applications has raised concerns about its disparate impact on minority groups and other vulnerable demographics. Motivated by a flurry of recent results highlighting bias in many automated decision making tasks such as facial recognition\citep{buolamwini2018gender} and criminal justice \citep{mehrabi2019survey}, researchers have begun focusing on mechanisms to ensure machine learning algorithms are \emph{fair} to all those affected. One of the challenges of fairness in an unsupervised learning context, compared to the supervised setting, is the lack of ground truth labels. Consequently, instead of enforcing approximately equal error rates across groups, fair clustering generally aims to ensure that composition of each cluster or each cluster's center, for settings like $k$-means and $k$-median clustering, fairly represent all groups \citep{chhabra2021overview}.

A common approach to fair clustering is to require each cluster to have a fair proportion of itself represented by each group (i.e. via balance \cite{chierichetti2017fair} or bounded representation \cite{ahmadian2019clustering}). However, this approach does not have the desired effect in settings where a group only gains a significant benefit from the cluster when they reach a minimum level of representation in their cluster. Consider a voting system where there are no constraints on the contiguity of how districts are designed and the goal is to design districts where voters are close together. Here, a proportionally fair clustering would assign a minority group that represents 30\% of the vote equally among each cluster. However, the minority group only gets a benefit (i.e. the ability to elect a candidate of their choice) if they have at least 50\% representation in the cluster. In this paper we introduce a new notion of fairness in clustering that tackles this problem. Specifically, we introduce \textit{minimum representation fairness} which requires each group to have a certain number of clusters where they cross a given minimum representation threshold (i.e. 50\% in the voting example).

Another real-world application of minimum representation fairness arises in entertainment segmentation. Consider the problem of grouping a set of media (e.g. television shows, songs) that need to be clustered into a set of segments (e.g. channels, playlists). A natural fairness consideration in designing these segments is ensuring that there is sufficient representation for different demographic groups. In these settings the benefit of the representation is only felt when a large percentage of the segment comes from a demographic group (i.e. so listeners can consistently watch or hear programming that speaks to them). This is even legislated in countries like Canada where, for example, Canadian television channels have to have at least 50\% Canadian programming \cite{crtc} Adding the additional criteria that the channels need to be cohesive (i.e. have similar genres and content), the problem of clustering content into channels subject to having sufficient group representation in enough channels fits into the minimum representation clustering framework.

Arguably the most popular algorithm for performing clustering is Lloyd's algorithm for $k$-means clustering \citep{jain1999data}. The algorithm is an iterative procedure that alternates between fixing cluster centers and assigning points to the closest clusters. The algorithm is guaranteed to return a local solution (i.e. no perturbation of the cluster centers around the solution leads to a better clustering cost). Unfortunately, naively using Lloyd's algorithm can lead to clusters that violate minimum representation fairness. Consider the following simple example from the adult dataset, which contains census data for 48842 individuals in 1994 \cite{dua2017uci}. Suppose we wanted to cluster these individuals into groups that represent different districts for a local committee and geographic contiguity was not a concern. A natural fairness criteria would be to ensure that there are a sufficient number of districts where minority groups (i.e. non-white in this dataset) have majority voting power. Despite the fact that approximately 15\% of the dataset is non-white (i.e. Black, Asian Pacific Islander, American Indian, or other), every cluster produced by Lloyd's algorithm is dominated by white members even when the number of clusters is as high as twenty. This highlights the need for a new approach to address fair minority representation.

In this paper we introduce a modified version of Lloyd's algorithm that ensures minimum representation fairness, henceforth referred to as MINIimum REpresentation fair Lloyd's algorithm (MiniReL for short). The key modification behind our approach is to replace the original greedy assignment step of assigning data points to its closest cluster center with an integer program (IP) that finds the minimum cost assignment while ensuring fairness. In contrast to the standard clustering setting, we show that finding a minimum cost clustering that respects minimum representation fairness is NP-Hard even when the cluster centers are already fixed. However, through numerical experiments, we also demonstrate that the IP approach is able to solve this problem to local optimality even for large datasets. We also present two computational approaches to improve the run-time including warm-starting our algorithm with the output of the standard Lloyd's algorithm and pre-assigning groups to specific clusters to help break symmetry and reduce the size of the IP that needs to be solved. We show empirically that our approach is able to construct fair clusters which have nearly the same clustering cost as those produced by Lloyd's algorithm.

\subsection{Minimum Representation Fair Clustering Problem}
The input to the standard clustering setting is a set of $n$ $m$-dimensional data points $\mathcal{X} = \{x^i \in \mathbbm{R}^m\}_{i=1}^n$. Note that assuming the data points to have real-valued features which is not a restrictive assumption in practice as categorical features can be converted to real-valued features through a one-hot encoding scheme. The goal of the clustering problem is to partition the data points into a set of $K$ clusters $\mathcal{C} = \{C_1, \dots, C_K \}$, where $C_k$ denotes the set of points belongs to cluster $k \in \mathcal{K} = \{1,\dots,K\}$, such that some measure of cluster quality is optimized. We focus on the popular $k$-means clustering metric which aims to find both a clustering and a set of centers $c_k\in \mathbbm R^m$ for each cluster so as to minimize the sum of the squared distance between each point and its cluster center. Formally:
$$
\min_{\mathcal{C}, c_1,\ldots,c_K} \sum_{k \in \mathcal{K}} \sum_{x^i \in C_k} \|x^i - c_k\|^2
$$
where $c_k$ denotes the center of each cluster $C_k$. In the absence of any additional constraints, for a given set of cluster centers the optimal cluster assignment is to simply assign each point to the nearest cluster center. Thus the problem can be viewed as an optimization over the choice of cluster centers. Note that the $k$-means clustering problem, even in the absence of fairness constraints, is both NP-hard and hard to approximate within a factor of $(1 + \epsilon)$ for any fixed $\epsilon>0$ \cite{jain1999data}. 

In the fair clustering setting, each data point belongs to a group $g\in \mathcal{G}$ (i.e. gender, race). Let $X_g$ be the set of data points belong to group $g$. Note that unlike other fair machine learning work, we do not assume that the groups form a partition of the data points. For instance, one instance might have groups corresponding to race and gender and a data point can belong to more than one group. The key intuition behind minimum representation fair clustering is that individuals belonging to a group only gain material benefit  if they have a minimum level of representation in their cluster. We denote this minimum representation threshold $\alpha \in (0,1]$, and define the associated notion of an $\alpha$-represented cluster as follows:

\begin{definition}[$\alpha$-represented Cluster]\label{def:alpha}
   A group $g\in \mathcal{G}$  is said to be $\alpha$-represented in a cluster $C_k$ if 
    $$
    | C_k \cap X_g | \geq \alpha |C_k|
    $$
    \end{definition}
    
Note that $\alpha$ represents the minimum threshold needed for a given group to receive benefit from a cluster and thus depends on the application. For instance, most voting systems require majority representation (i.e. $\alpha = 0.5$). Our framework also allows for $\alpha$ to be group-dependent (i.e. $\alpha_g$ for each group $g$), however in most applications of interest $\alpha$ is a fixed threshold regardless of group. For a given clustering $\mathcal{C}$, group $g$, and $\alpha$, let $\Lambda(\mathcal{C}, X_g, \alpha)$ be the number of clusters $\alpha$-represented by group $g$. In minimum representation fairness, each group $g$ has a parameter $\beta_g$ that specifies a minimum number of clusters that should be $\alpha$-represented by that group.

\begin{definition}[Minimum representation fairness]
A given clustering $\mathcal{C} = \{C_1, \dots, C_K\}$ is said to be an ($\alpha,\boldsymbol{\beta}$)-minimum representation fair clustering if for every group $g \in {\mathcal{G}}$:
$$
\Lambda(\mathcal{C}, X_g, \alpha) \geq \beta_{g}
$$
for a given $\boldsymbol{\beta} = \{\beta_g \in \mathbb{Z}^+\}_{g \in \mathcal{G}}$.
\end{definition}

The definition of minimum representation fairness is flexible enough that the choice of  $\boldsymbol{\beta}$ can and should be specialized to each application as well as the choice of $\alpha$. In the remainder of the paper we explore two different natural choices for $\boldsymbol{\beta}$ that mirror fairness definitions in the fair classification literature. The first sets $\beta_g$ to be equal for all groups, which we denote cluster statistical parity. 
\begin{definition}[Cluster Statistical Parity]
A given clustering is said to meet cluster statistical parity if it is a minimum representation fair clustering for a given $\alpha$ and the following $\beta$:
$$
\beta_g = \Big\lfloor \frac{1}{|\mathcal{G}|} \big\lfloor \alpha^{-1} \big \rfloor K \Big \rfloor \quad \forall g \in \mathcal{G}
$$
\end{definition}
The second sets $\beta_g$ to be proportional to the size of the group, which we denote cluster equality of opportunity. 
\begin{definition}[Cluster Equality of Opportunity]
A given clustering is said to meet cluster equality of opportunity if it is a minimum representation fair clustering for a given $\alpha$ and the following $\beta$:
$$
\beta_g = \Big \lfloor \frac{|X_g|}{n} \big \lfloor \alpha^{-1} \big\rfloor K \Big \rfloor \quad \forall g \in \mathcal{G}
$$
\end{definition}
In both definitions $\lfloor \alpha^{-1} \rfloor$ corresponds to the maximum number of groups that can be $\alpha$-represented in a cluster.
Combining the standard $k$-means problem with minimum representation fairness criteria gives the following formal optimization problem:

\begin{definition}[Minimum representation fair $k$-means problem]
For a given $\alpha \in (0,1]$ and $\boldsymbol{\beta} = \{\beta_g \in \mathbb{Z}^+\}_{g \in \mathcal{G}}$, the minimum representation fair k-means problem is:
$$
\min_{\mathcal{C}, c_1,\ldots,c_K} \sum_{k \in \mathcal{K}}\sum_{x^i \in C_k} \|x^i - c_k\|^2
\quad \textbf{s.t.} \quad 
\Lambda(\mathcal{C}, X_g, \alpha) \geq \beta_{g} ~~\forall g \in \mathcal{G} 
$$
\end{definition}

An important difference between the  fair  and the standard versions of the $k$-means clustering problem is that greedily assigning data points to their closest cluster center may no longer be optimal for the fair version (i.e. assigning a data point to a farther cluster center may be necessary to meet the fairness criteria). Thus the problem can no longer be viewed simply as an optimization problem over cluster centers.

\subsection{Related Work}
A recent flurry of work in fair clustering has given rise to a number of different notions of fairness. One broad line of research, started by the seminar work of Chierichetti et al. \cite{chierichetti2017fair}, puts constraints on the \textit{proportion} of each cluster that comes from different groups. This can be in the form of balance \cite{chierichetti2017fair, bera2019, schmidt2018fair, bercea2018cost, backurs2019scalable, kleindessner2019guarantees, ahmadian2020fair, bohm2020fair, chhabra2020fair, liu2021stochastic, ziko2021variational, le2021fair} which ensures each group has relatively equal representation, or a group specific proportion such as the bounded representation criteria \cite{ahmadian2019clustering, bera2019, ahmadian2020fair, schmidt2018fair, esmaeili2020probabilistic, jia2020fair, huang2019coresets, bandyapadhyay2020coresets, harb2020kfc} or maximum fairness cost \cite{chhabra2020fair}. Minimum representation fairness bares a resemblance to this line of work as it puts a constraint on the proportion a group in a cluster, however instead of constraining a fixed proportion across all clusters it looks holistically across all clusters and ensures that threshold is met in a baseline number of clusters.

Another line of work tries to minimize the \textit{worst case average clustering cost} (i.e. $k$-means cost) over all the groups, called social fairness \cite{ghadiri2021socially, abbasi2021fair, makarychev2021approximation, goyal2021tight}. Most similar to our algorithmic approach is the Fair Lloyd algorithm introduced in \cite{ghadiri2021socially}. They also present a modified version of Lloyd's algorithm that converges to a local optimum. However their approach requires a modified center computation step that can be done in polynomial time. Conversely, our problem requires a modified cluster assignment step that is NP-hard which we solve via integer programming.

Most similar to minimum representation fairness is diversity-aware fairness introduced in \cite{thejaswi2021diversity} and the related notion of fair summarization \cite{kleindessner2019fair, chiplunkar2020solve, jones2020fair}. These notions of fairness require that amongst all the cluster centers selected, a minimum number comes from each group. Minimum fairness representation differs in that our criteria is not tied to the group membership of the cluster center selected but the proportion of each group in a given cluster. Our notion of fairness makes more sense in settings where the center cannot be prescribed directly, but is only a function of its composition (i.e. in voting where members of a 'cluster' elect an official).    

There is also a long line of research that looks at fairness in the context of gerrymandering \citep{kueng2019fair, gurnee2021fairmandering, benade2022political, levin2019automated, mehrotra1998optimization, ricca2013political}. While our notion of fairness shares some similarity with different notions of fairness in gerrymandering, the gerrymandering problem places different constraints on the construction of the clusters such as contiguity. Consequently the algorithmic approaches to tackle gerrymandering generally require more computationally intensive optimization procedures that do not readily transfer to the machine learning setting.
    
\subsection{Main Contributions}
We summarize our main contributions as follows:
\begin{itemize}
\item We introduce a novel definition of fairness for clustering called minimum representation fairness, which requires that a specified number of clusters should have at least $\alpha$ percent members from a given group.
\item We show that our given definition of fairness encompasses analogs to fair classification metrics such as statistical parity and equality of opportunity.
\item We formulate the problem of finding a minimum representation k-means clustering in a mixed integer optimization framework, and introduce a new heuristic algorithm MiniReL, based on Lloyd's algorithm, to find a local optimum.
\item We show that incorporating minimum representation fairness into Lloyd's algorithm leads to a NP-Hard sub-problem. To tackle this issue, we develop computational techniques that make the our approach tractable even for large datasets.
\item We present numerical results to demonstrate that MiniReL is able to construct minimum representation fair clusterings with only a modest increase in run-time and practically with no loss in clustering quality compared to the standard k-means clustering algorithm.  
\end{itemize}

The remainder of the paper is organized as follows. In section \ref{sec:mio} we present a mixed integer optimization formulation for the minimum representation fair clustering problem and introduce MiniReL, a variant of Lloyd's algorithm to handle the fairness constraints. In section \ref{sec:scaling} we introduce computational approaches to help our algorithm scale to large datasets. Finally section \ref{sec:exp} presents a numerical study of MiniReL compared to the standard k-means algorithm.

\section{Mixed Integer Optimization Framework} \label{sec:mio}
We start by formulating the minimum representation fair clustering problem as a mixed-integer program with a non-linear objective. We use binary variable $z_{ik}$ to denote if data point $x^i$ is assigned to cluster $k$, and variable $c_k \in \mathbb{R}^d$ to denote the center of cluster $k$. Let $y_{gk}$ be the binary variable indicating whether group $g$ is $\alpha$-represented in cluster $k$. Finally, let $\mathcal{W} \subseteq \mathcal{G} \times \mathcal{K}$ be the set of allowable clusters that can be $\alpha$-represented by each group. In most applications $\mathcal{W}$ will be equal to $\mathcal{G} \times \mathcal{K}$, however in some applications it is beneficial to restrict this set. For instance, we show in Section \ref{sec:scaling} that pre-fixing groups to specific clusters (i.e specifying cluster $i$ must be $\alpha$-represented by group $g$) can help break symmetry in the IP model and dramatically speedup the runtime of the algorithm.

We can now formulate the minimum representation fair clustering problem as follows:
\begin{align}
	\hskip1.5cm\textbf{min}~~&& \sum_{x^i \in {\mathcal X}} \sum_{k \in \mathcal{K}} \|x^i - c_k\|_2^2 z_{ik}\label{obj:mip}\\
	\textbf{s.t.}~~&& \sum_{k \in \mathcal{K}}  z_{ik} &= 1 ~~&&\forall x^i \in {\mathcal X} \label{const:cluster_assignment}\\
	&& \sum_{x^i \in X_g} z_{ik} +  M (1 - y_{gk})&\geq \alpha \sum_{x^i \in {\mathcal X}} z_{ik}    ~~ &&\forall (g,k) \in {\mathcal W}\label{const:dominance}\\
	&& \sum_{k \in \mathcal{K}: (g,k) \in \mathcal{W}} y_{gk} &\geq \beta_g &&\forall g \in {\mathcal G} \label{const:num_dominate}\\ 
 && l\leq \sum_{x^i \in {\mathcal X}} z_{ik} &\leq u &&\forall k \in {\mathcal K} \label{const:cardinality} \\
	&&z_{ik}, y_{gk} &\in \{0,1\} &&\forall x^i \in \mathcal{X}, (g,k) \in \mathcal{W} \label{const:binary}
\end{align}

The objective \eqref{obj:mip} is to minimize the sum of squares cost of the clustering. Constraint \eqref{const:cluster_assignment} ensures that each data point is assigned to exactly one cluster. Constraint \eqref{const:dominance} tracks whether a cluster $k$ is $\alpha$-represented by a group $g$, and includes a big-$M$ which can be set to $\alpha n$. Finally, constraint \eqref{const:num_dominate} tracks that each group $g$ is $\alpha$-represented  in at least $\beta_g$ clusters. In many applications of interest, it might also be worthwhile to add a constraint on the size of the clusters to ensure that each cluster has a  minimum/maximum number of data points. Constraint \eqref{const:cardinality} captures this notion of a cardinality constraint where $l$ and $u$ represent the lower and upper bound for the cardinality of each cluster respectively. Note that in cases where the cardinality constraint is used, the big-$M$ in constraint \eqref{const:dominance} can be reduced to $\alpha u$. For all our experiments we set $l = 1$ to ensure that exactly $k$ clusters are returned by the algorithm. Adding a lower bound also ensures that each group is $\alpha$-represented in non-trivial clusters. Note that every group would be trivially $\alpha$-represented in an empty cluster according to Definition \ref{def:alpha} but would provide little practical use.

\subsection{MiniReL Algorithm}

Solving the optimization problem outlined in the preceding section to optimality is computationally challenging as it is an integer optimization problem with a non-linear objective. Instead of solving this problem directly, we take an approach similar to Lloyd's algorithm and alternate between adjusting cluster centers and assigning data points to clusters to converge to a local optimum. Given a fixed set of cluster assignments (i.e. when variables $z$ are fixed in \eqref{obj:mip}-\eqref{const:binary}) the optimal choice of $c_k$ is the mean value of data points assigned to $C_k$.
However,  when the cluster centers are given, the assignment problem is non-trivial due to the fairness constraints. For a fixed set of cluster centers $c_k$ we denote the problem \eqref{obj:mip}-\eqref{const:binary} the \textit{fair assignment problem}. Note that unlike the full formulation, this is a linear integer program. In Lloyd's algorithm the assignment stage (i.e. solving \eqref{obj:mip}-\eqref{const:cluster_assignment}) can be done greedily in polynomial time. However, the introduction of the fairness and cardinality constraints makes this no longer possible. The following result shows that even finding a feasible solution to the fair assignment problem is NP-Complete (proof can be found in Appendix \ref{sec:proof_np}).

\begin{theorem} \label{thm:fair_assign_np}
Finding a feasible solution to the fair assignment problem is NP-Complete.
\end{theorem}

Given that the fair assignment problem is NP-Complete, we solve it to optimality using integer programming. While integer programming tends not to scale well to large problems, in practice we observed the fair assignment problem to be computationally tractable for even datasets with thousands of data points. In Section \ref{sec:scaling} we also discuss some computational approaches to improve the run time for larger datasets. We denote our modified version of Lloyd's algorithm the Minimum Representation Fair Lloyd's Algorithm (MiniReL), which is summarized in Algorithm \ref{alg:minrepfairlloyd}. 

\begin{algorithm}[tb]
\caption{Minimum Representation Fair Lloyd's Algorithm (MiniReL)}
\label{alg:minrepfairlloyd}
\begin{flushleft}

\textbf{Input}: Data $\mathcal{X}$, Number of clusters $K$, Fairness parameters $\beta$, $\alpha$, $\mathcal{W}$, Iteration limit $L$

\textbf{Output}: Cluster assignments $\{C_k\}_{k \in \mathcal{K}}$ 
\end{flushleft}

\begin{algorithmic}[1] 
\STATE Initialize $\{c_k\}_{k \in \mathcal{K}}$  (i.e. by uniformly at random sampling $K$ data points).
\FOR{$l=1,2,\dots, L$}
\STATE /* Update Cluster Assignments */
\STATE Solve fair assignment problem \eqref{obj:mip}-\eqref{const:binary} to get \textbf{z}* for fixed cluster centers $\{c_k\}_{k \in \mathcal{K}}$ \\
\STATE /* Update Cluster Centers */
\FOR{$k=1,2,\dots, K$}
\STATE Set  $N_k = \sum_{x^i \in \mathcal{X}} z_{ik}$, $c_k = \frac{1}{N_k} \sum_{x^i \in \mathcal{X}} z_{ik} x^i$. \\
\ENDFOR
\ENDFOR
\STATE Set $C_k = \{x^i\in \mathcal{X} : z_{ik} = 1\} \quad \forall k \in \mathcal{K}$
\STATE \textbf{return} $\{C_k\}_{k=1}^K$
\end{algorithmic}
\end{algorithm}

A natural question is whether MiniReL maintains the convergence guarantees of Lloyd's algorithm which is guaranteed to converge in finite time to a local optimum. When discussing a local optimum it is important to formally define a local neighborhood for a solution. Traditionally, a clustering via Lloyd's algorithm is defined by the location of the centers (i.e. a perturbation of the centers can change the cluster assignment). This leads to settings where multiple partitions for a given set of centers need to be tested (see \cite{ghadiri2021socially} for a discussion) to ensure local optimality. However in the minimum representation fairness setting a data point may be assigned to a cluster that is not its closest center and therefore a local change to a cluster center should not change the cluster assignment. In this setting, we define a local change as any perturbation to a cluster center, and any individual change to cluster assignment (i.e. moving a data point from one cluster to another). With this notion of local neighborhood, the following result shows that the MiniReL also converges to a local optimum in finite time (proof can be found in Appendix \ref{sec:proof_convergence}).

\begin{theorem} \label{thm:finite_convergence} 
MiniReL converges to a local optimum in finite time.
\end{theorem}

\section{Scaling MiniReL} \label{sec:scaling}
In MiniReL, the computational bottleneck is solving the fair assignment problem which is an IP. To improve the run-time of MiniReL, we introduce two key computational approaches that reduce the number of fair assignment problems that need to be solved, and improve the speed at which they can be solved.

\subsection{Warm-Starting with K-means} \label{sec:warm_start}
To reduce the number of iterations needed to converge in MiniReL, we warm-start the initial cluster centers with the final centers of the conventional Lloyd's algorithm. The key intuition behind this approach is that it allows us to leverage the polynomial time assignment problem for the majority of iterations, and only requires solving the fair assignment problem to adjust a local unfair optimum to a fair one.

\begin{figure}[t]
    \centering
    \begin{subfigure}
      \centering
      \includegraphics[width=\textwidth]{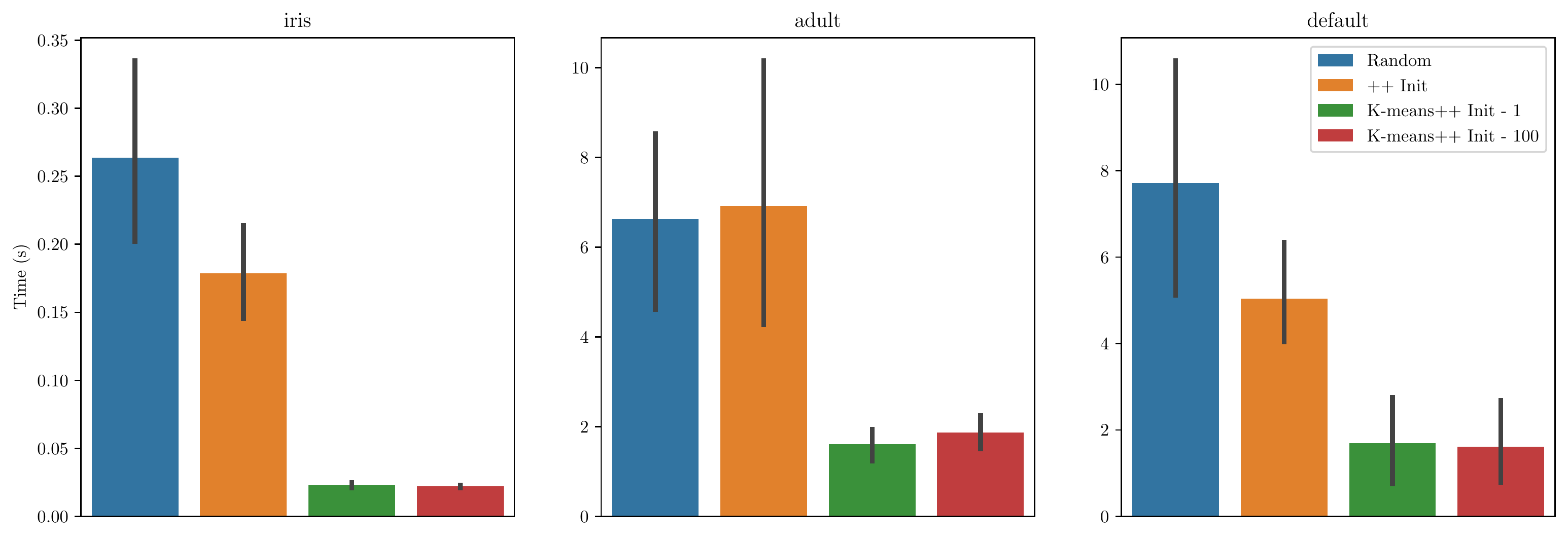}
    \end{subfigure}
  \caption {\label{fig:init_time} Average computation time in seconds over 10 random seeds for MiniReL with different initialization schemes. Bars indicate standard error.
  }
\end{figure}

To incorporate warm-starting into MiniReL, we replace step 1 in Algorithm \eqref{alg:minrepfairlloyd} with centers generated from running Lloyd's algorithm with the $k$-means++ initialization \cite{arthur2006k}. We benchmark this approach against two baselines: $(i)$ randomly sampling the center points, and $(ii)$ using the $k$-means++ initialization scheme without running Lloyd's algorithm afterwards. We also compare using the $k$-means warm-start with 1 initialization and 100 initializations. Figure \ref{fig:init_time} shows the impact of these initialization schemes on the total computation time including time to perform the initialization. Each initialization scheme was tested on three datasets. For each dataset we  randomly sub-sample 2000 data points (if $n > 2000$), and re-run MiniReL with 10 random seeds. The results show that using Lloyd's algorithm to warm-start MiniReL can lead to a large reduction in computation time, even taking into account the cost of running the initialization. However, there are diminishing returns. Namely running 100 different initialization for $k$-means and selecting the best leads to slightly larger overall run-times. Additional results on the impact of these initialization schemes on the number of iterations until convergence and the clustering cost of the final solution are included in Appendix \ref{sec:exp_inits} but confirm that warm-starting MiniReL with $k$-means leads to better solutions in less time.

\subsection{Pre-fixing Group Assignment}
One computational shortcoming of the fair assignment IP model is the use of big-M constraints which are well known to lead to weak continuous relaxations and by extension longer computation times. However, their use in the IP model is simply to track which groups are  $\alpha$-represented in  which clusters.  One approach to avoid the need for this tracking is to simply pre-fix which groups need to be represented by which  clusters (i.e. akin to fixing the $y_{gk}$ variable). This removes the need for the $y$ variables and the associated big-M constraints and breaks some symmetry in the IP (i.e. removes permutations of an optimal solution that are also feasible), dramatically improving the problem's computation time. In problems where only a single group can be $\alpha$-represented by a cluster (i.e. a data point can only be part of one group, and $\alpha > 0.5$), doing this pre-fixing preserves the optimal solution to the full problem. However, in more complicated settings (i.e. multiple intersecting groups, $\alpha \leq 0.5$) pre-fixing may remove the optimal solution and thus simply represents a heuristic for improving run-time. It is worth noting that the MiniReL algorithm is itself a heuristic, and thus the pre-fixing scheme has ambiguous effects on the quality of the solution (i.e. may force the algorithm to converge to a better local optimum). 

\begin{figure}[t]
    \centering
    \begin{subfigure}
      \centering
      \includegraphics[width=\textwidth]{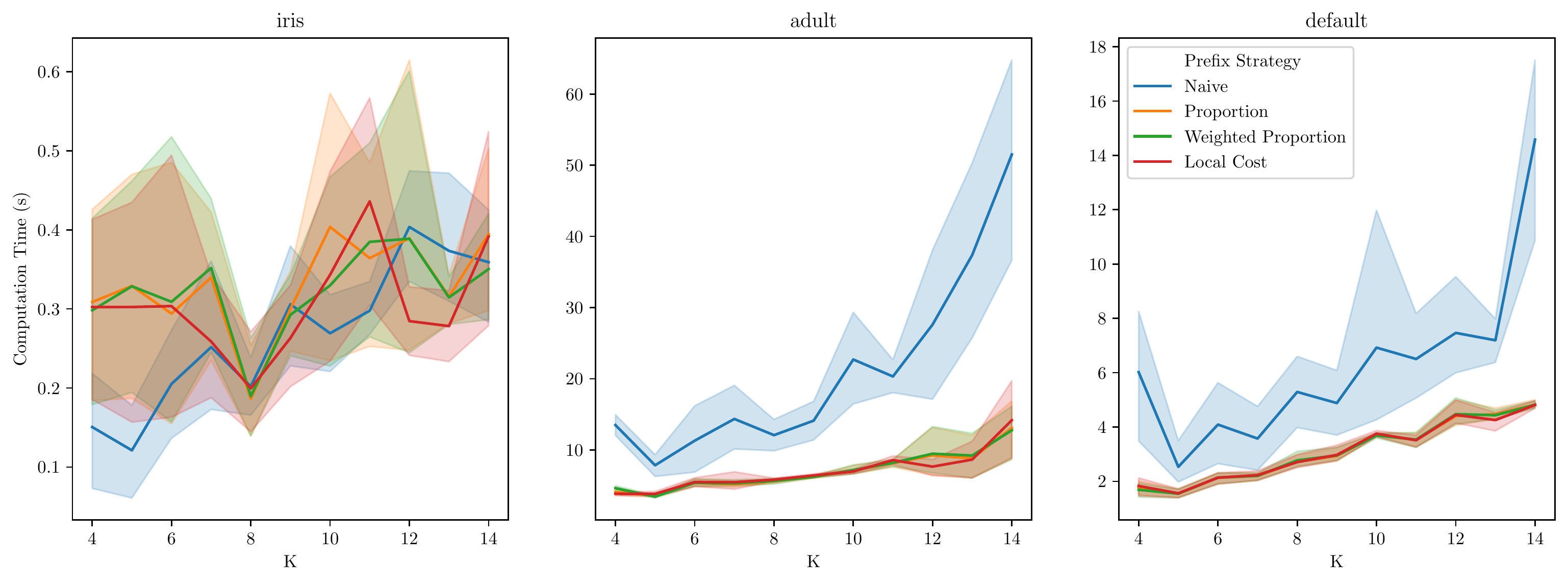}
    \end{subfigure}
  \caption {\label{fig:prefix_time} Impact of pre-fixing strategy on the run-time of the MiniReL algorithm. Proportion, Weighted Proportion, and Local Cost are pre-fixing strategies that use the pre-fix IP model with different objectives. Naive denotes random assignments of groups to clusters. Results averaged over 10 random seeds, widths indicate standard error.
  }
\end{figure}
Given a warm-started initialization to MiniReL (i.e. from $k$-means as outlined above), a natural question is how to perform the pre-fixing so as to minimize the number of iterations or computation time needed to find a solution. We can formulate the problem of finding the best pre-fixing as a small integer program. Let $x_{g,k}$ be a binary variable indicating whether group $g$ is $\alpha$-represented in cluster $k$. Let $c_{gk}$ be the cost associated with $\alpha$-representing group $g$ in cluster $k$. We experiment with three different choices of cost function:
\begin{itemize}
    \item \textit{Proportion}: Set the cost to the proportion of the cluster that needs to be changed for $g$ to be $\alpha$-represented in cluster $k$: $$c_{gk} = \max(\alpha - p_{gk},0)$$ where $p_{gk}$ is the current proportion of cluster $k$ that belongs to group $g$.
    \item \textit{Weighted Proportion}: Set the cost to the proportion weighted by the size of the cluster: $$ c_{gk} = |C_k|\max(\alpha - p_{gk},0).$$
    \item \textit{Local Cost}: Set the cost to the myopic $k$-means loss to move the necessary number of points, $q$, for $g$ to be $\alpha$-represented in cluster $k$ (i.e. $q + p_{gk}|C_k| \geq \alpha(q + |C_k|)$). Formally, $q = \lceil \frac{\max(0,\alpha-p_{gk})}{1 - \alpha} |C_k| \rceil$. The local cost is:
    $$
    c_{gk} = \min_{X \subset X_g \setminus C_K: |X| = q} \sum_{x \in X} \|x - c_k\|_2^2
    $$
\end{itemize}

\newpage

\begin{figure}[!b]
    \centering
    \begin{subfigure}
      \centering
      \includegraphics[width=\textwidth]{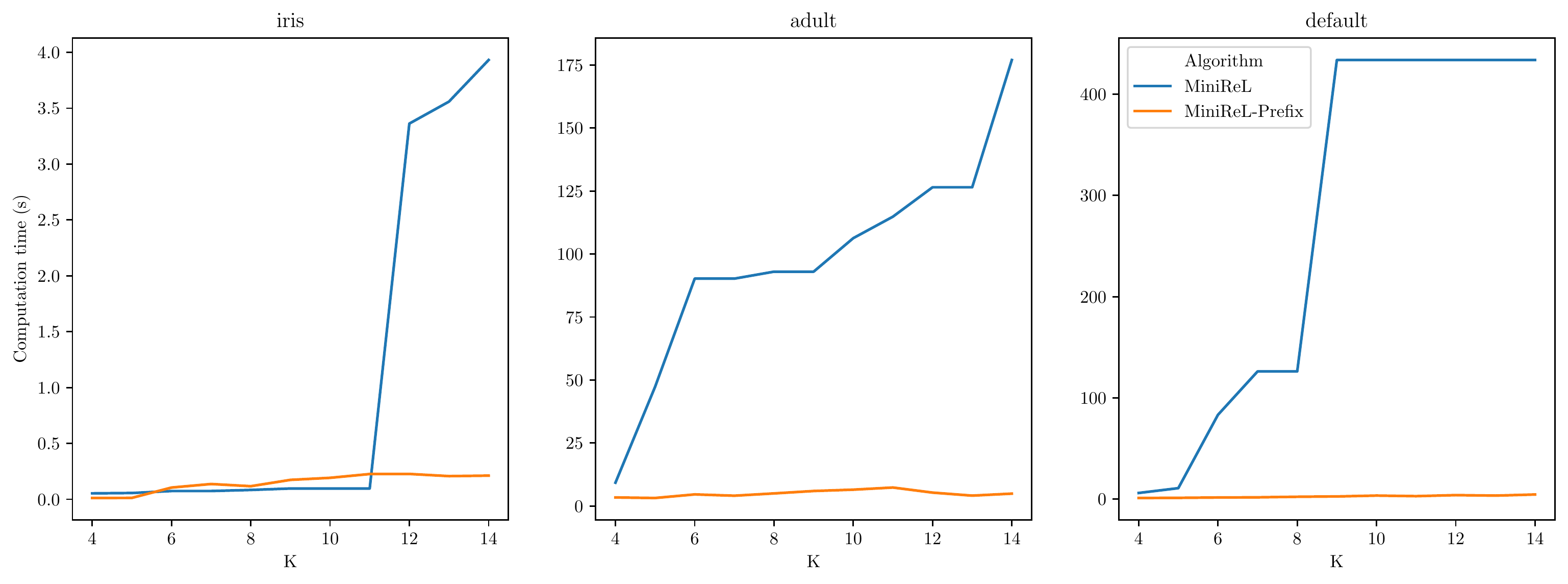}
    \end{subfigure}
  \caption {\label{fig:prefix_timesave} Impact of pre-fixing on runtime of the MiniReL Algorithm.
  }
\end{figure}
We can now formulate an IP to perform the pre-fix assignment as follows:

\begin{align}
	\hskip1.5cm\textbf{min}~~&& \sum_{(g,k) \in {\mathcal W}} c_{gk} x_{gk}\label{obj:prefix}\\
	\textbf{s.t.}~~&& \sum_{k \in \mathcal{K}:(g,k) \in \mathcal{W}}  x_{gk} &\geq \beta_g ~~&&\forall g \in {\mathcal G} \label{const:beta_const}\\
	&& \sum_{g \in \mathcal{G}}x_{gk} &\leq \Big\lfloor \frac{1}{\alpha} \Big\rfloor ~~ &&\forall k \in \mathcal{K} \label{const:cluster_prefix_cap}\\
	&& x_{gk} &\in \{0,1\} ~~ &&\forall (g,k) \in \mathcal{W} \label{const:prefix_binary}
\end{align}

The objective \eqref{obj:prefix} is simply to minimize the cost of pre-fixing. Constraint \eqref{const:beta_const} ensures enough clusters are allocated to each group to meet the minimum representation fairness constraint. Finally constraint \eqref{const:cluster_prefix_cap} ensures no cluster is assigned more groups than can simultaneously be $\alpha$-represented by it. 
Note that this IP is small and only scales with $\mathcal{W} \leq |\mathcal{K}| \times |\mathcal{G}|$. In practice we found that solving this IP took under 1 second for all instances tested in this paper.

Figure \ref{fig:prefix_time} shows the impact of different choices of the objective in IP model \eqref{obj:prefix}-\eqref{const:prefix_binary} on run-time for MiniReL on three different datasets. For all these experiments we ran 10 trials with different random initial seeds, and warm-start the algorithm with the standard Lloyd's algorithm. We benchmark using the IP model to perform pre-fixing with naively pre-fixing the group cluster assignments (i.e. random assignment). In the small 150 data point iris dataset, the pre-fixing scheme has little impact on the total run-time of the algorithm as the overhead of running the IP model outweighs any time savings from a reduced number of iterations. However, for larger datasets (i.e. adult and default which both have over 30K data points), using the IP model to perform pre-fixing outperforms the naive approach - leading to as large as a 3x speed-up. However, there is a relatively small difference in performance between the three choices for the objective, with the local cost objective reducing the speed by approximately $0.1\%$ compared to the other two when averaged across all three datasets. For the remainder of this paper, all MiniReL results use prefixing with the local cost objective. Additional results highlighting the impact of the pre-fixing objective on the number of iterations until convergence and the clustering cost of the final solution are included in Appendix \ref{sec:exp_prefix}.

The speedups from pre-fixing group cluster assignments is more pronounced when compared to running MiniReL with the full IP model for fair assignment. Figure \ref{fig:prefix_timesave} compares the run-time of the two algorithms and shows that MiniReL with prefixing can lead to over a 400x speedup  in some instances. Results in Appendix \ref{sec:exp_prefix} also confirm that these speedups come at no cost to the quality of the final solution.

\section{Numerical Results} \label{sec:exp}
\begin{figure}
    \centering
    \begin{subfigure}
      \centering
      \includegraphics[width=\textwidth]{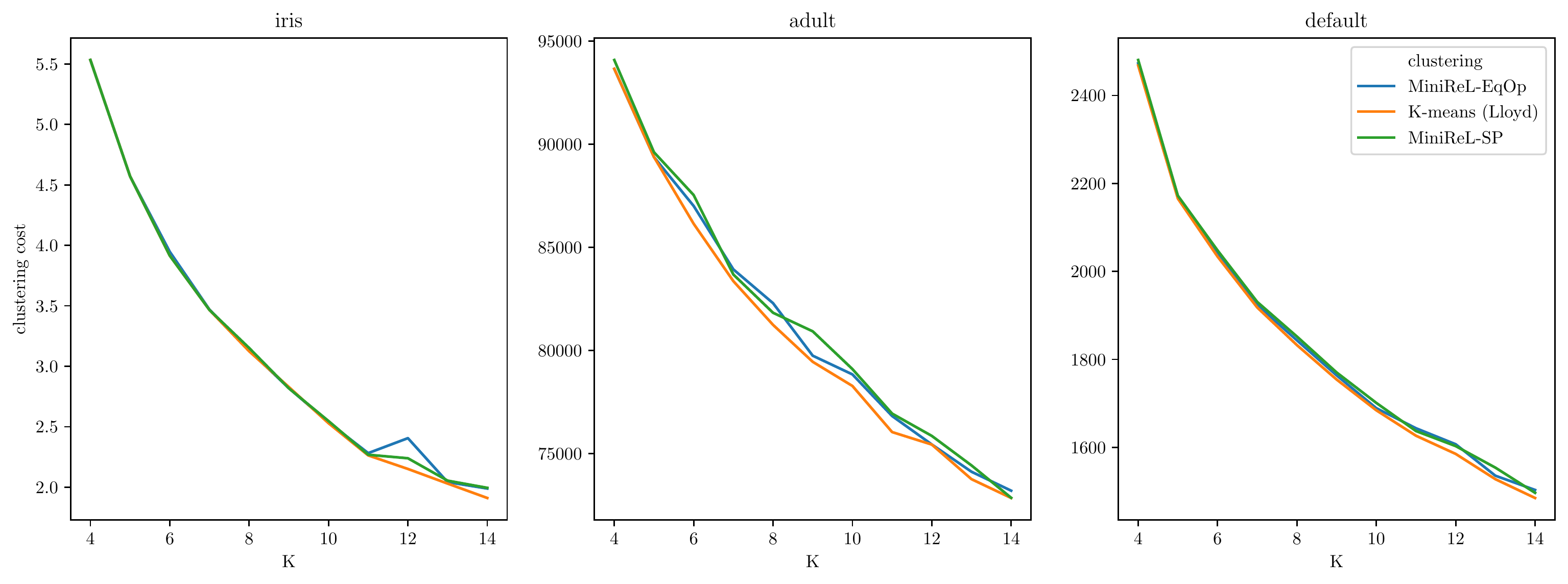}
    \end{subfigure}
  \caption {\label{fig:cluster_cost_joint} $k$-means clustering cost of Lloyd's algorithm and MiniReL.
  }
\end{figure}
\begin{figure}
    \centering
    \begin{subfigure}
      \centering
      \includegraphics[width=\textwidth]{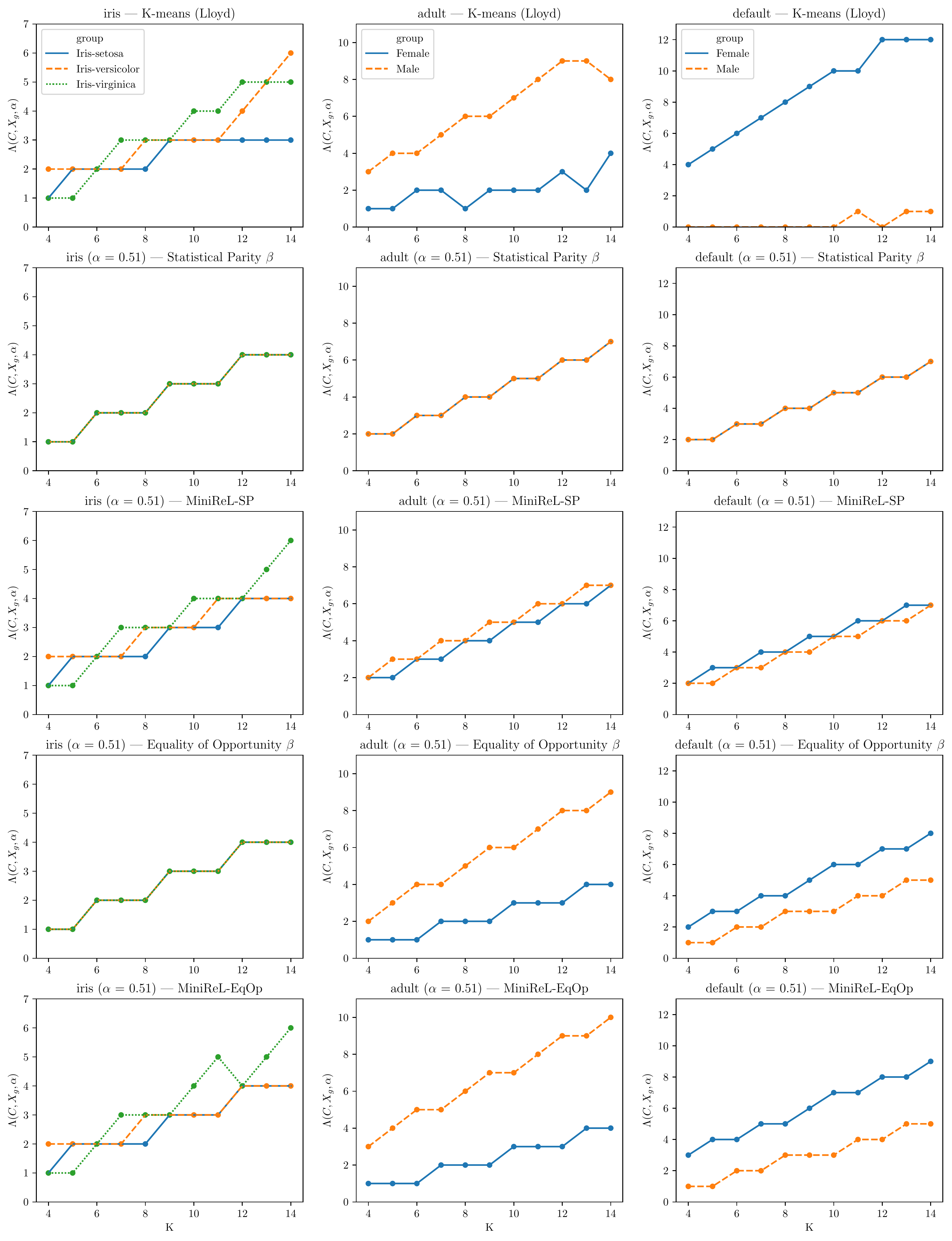}
    \end{subfigure}
  \caption {\label{fig:fairness_joint} Number of $\alpha$-represented clusters for $k$-means and MiniReL algorithms. Rows 2 and 4 indicate $\beta$ setting for cluster \textit{statistical parity} and cluster \textit{equality of opportunity } respectively.
  }
\end{figure}

To benchmark our approach, we evaluate it on three datasets from the UCI machine learning repository \cite{dua2017uci} that have been used in recent work in fair clustering: iris ($n=150$, $m=4$), adult ($n=48842$, $m=14$), and default ($n=30000$, $m=24$).For each dataset we use one sensitive feature to represent group membership - namely species for Iris, and gender for both adult and default. For all datasets we normalize all real-valued features to be between $[0,1]$ and convert all categorical features to be real-valued via a one-hot encoding scheme. For datasets that were originally used for supervised learning we remove the target variable, and do not use the sensitive attribute as a feature for the clustering itself. 

We compare MiniReL against the standard Lloyd's algorithm using the implementation available in scikit-learn \cite{scikit-learn} with a $k$-means++ initialization. For all $k$-means results we re-run the algorithm with 10 different random seeds and report the result with the best $k$-means cost.  We implemented MiniReL in Python with Gurobi 10.0 \cite{gurobi} for solving all IPs. To initialize the cluster centers we use the warm start scheme outlined in Section \ref{sec:scaling}. 
We ran MiniReL with $\boldsymbol{\beta}$ set for both cluster statistical parity (MiniReL-SP) and cluster equality of opportunity (MiniReL-EqOp). For the following experiments we set $\alpha = 0.51$ to represent majority representation in a cluster.

Figure \ref{fig:fairness_joint} show the number of $\alpha$-represented clusters for each group using both $k$-means and MiniReL. Rows 2 and 4 of both plots show the fair baseline (i.e. the settings for $\beta$). Across all three datasets we can see that $k$-means can lead to outcomes that violate minimum representation fairness constraints. This is most stark in the default dataset where there is as much as an 11 cluster gap between the two groups despite having similar proportions in the dataset (60\% and 40\% for females and males respectively). In contrast, the MiniReL algorithm is able to generate fair clusters under both notions of fairness and for all three datasets. A natural question is whether this fairness comes at cost to the quality of the clustering. Figure \ref{fig:cluster_cost_joint} shows the $k$-means clustering cost for Lloyd's algorithm and MiniReL under both definitions of fairness. Although there is a small increase in the cost when using MiniReL the overall cost closely matches that of the standard $k$-means algorithm showing that we can gain fairness at practically no additional cost to cluster quality. However, running MiniReL does come at a cost in terms of computation time. Table \ref{tab:timing} shows the average computation time in seconds for both algorithms. As expected, the harder assignment problem in MiniReL leads to higher overall computation times than the standard Lloyd's algorithm. However, it is still able to solve large problems in under 30 seconds demonstrating that the approach is still of practical use.

\begin{table}
\footnotesize
\caption{\label{tab:timing} Computation time in seconds over 10 random seeds (standard deviation in parentheses). Final row includes average computation time over all random seeds and setting of K for each dataset.}.
\begin{tabular}{l|cc|cc|cc}
\toprule
 & \multicolumn{2}{c|}{iris} & \multicolumn{2}{c|}{adult} & \multicolumn{2}{c}{default} \\
K & K-means (Lloyd) &     MiniReL & K-means (Lloyd) &      MiniReL & K-means (Lloyd) &      MiniReL \\
\midrule
4    &       0.0 (0.0) &  0.3 (0.13) &      0.6 (0.03) &  13.4 (0.25) &      0.3 (0.02) &  10.8 (0.18) \\
5    &       0.0 (0.0) &  0.4 (0.13) &      0.7 (0.04) &  14.3 (0.37) &      0.4 (0.03) &  11.2 (0.14) \\
6    &       0.0 (0.0) &  0.4 (0.24) &      0.8 (0.04) &  15.8 (1.08) &      0.5 (0.05) &  12.0 (0.33) \\
7    &       0.0 (0.0) &  0.3 (0.11) &      0.9 (0.04) &  15.8 (1.03) &      0.6 (0.05) &  12.3 (0.35) \\
8    &       0.0 (0.0) &  0.3 (0.05) &      1.0 (0.05) &  17.2 (0.54) &      0.8 (0.09) &  13.3 (0.29) \\
9    &       0.0 (0.0) &  0.3 (0.07) &      1.1 (0.06) &  17.8 (0.91) &      0.8 (0.08) &  13.5 (0.37) \\
10   &       0.0 (0.0) &  0.4 (0.09) &      1.3 (0.04) &   19.1 (0.9) &      0.9 (0.07) &  14.8 (0.29) \\
11   &       0.0 (0.0) &   0.4 (0.1) &      1.4 (0.04) &  19.6 (1.16) &      1.0 (0.06) &   14.4 (0.3) \\
12   &       0.0 (0.0) &  0.4 (0.17) &      1.4 (0.04) &  21.4 (3.77) &      1.1 (0.11) &  15.6 (0.55) \\
13   &       0.0 (0.0) &  0.4 (0.09) &      1.5 (0.05) &  20.9 (3.01) &      1.3 (0.18) &  16.0 (0.36) \\
14   &       0.0 (0.0) &  0.4 (0.09) &      1.7 (0.07) &  22.8 (3.02) &      1.5 (0.21) &  16.7 (0.35) \\
15   &      0.0 (0.01) &  0.5 (0.08) &      1.8 (0.08) &   21.4 (1.7) &      1.5 (0.11) &  17.5 (1.68) \\ \midrule
\textbf{Overall} &      \textbf{0.0 (0.01)} &  \textbf{0.4 (0.13)} &      \textbf{1.2 (0.39)} &  \textbf{18.3 (3.43)} &      \textbf{0.9 (0.41)} &  \textbf{14.0 (2.17)} \\
\bottomrule
\end{tabular}
\end{table}

\section{Conclusion}
In this paper we introduce a novel definition of group fairness for clustering that ensures each group achieves a minimum level of representation in a specified number of clusters. This definition is a natural fit for a number of real world examples, such as voting and entertainment segmentation. Unfortunately, the popular Lloyd's algorithm for $k$-means clustering results in unfair outcomes. To create fair clusters we introduce a modified version of Lloyd's algorithm called MiniReL that includes an NP-hard fair assignment problem. To solve the fair assignment problem we use integer programming, and present some computational tools to improve the run-time of the approach. Unfortunately, solving the integer program remains a computational bottleneck of the approach and an important area of future research is to design more efficient algorithms for the problem or heuristics that can guarantee approximate fairness in polynomial time. Nevertheless we note that our approach is able to solve problems of practical interest, including datasets with thousands of data points, and provides a mechanism to design fair clusters when Lloyd's algorithm fails.

\bibliographystyle{plain}
\bibliography{main}

\appendix

\section{Proof of Theorem \ref{thm:fair_assign_np} } \label{sec:proof_np}
\begin{proof}
Given an instance of a 3-SAT problem, one of Karp's 21 NP-Complete problems, we construct an instance of the fair assignment problem as follows. We start with a 3-SAT problem with $n$ variables and $m$ clauses $K_1, \dots, K_m$. Each clause $K_i$ takes the form $v_{i1} \vee v_{i2} \vee v_{i3}$ where $v_{ij}$ is either one of the original variables or its negation. 

To construct the instance of the fair assignment problem, start by creating two new data points $x_{v_i}, x_{\bar{v}_i}$, corresponding to each original variable $v_i$ and its negation respectively. We construct two clusters $C_1$ and $C_2$ that each variable can be assigned to. For each original variable $v_i$ we create one group $g_i = \{x_{v_i}, x_{\bar{v}_i}\}$ that can be $\alpha$-represented in either cluster (i.e. $(g_i, 1), (g_i,2) \in \mathcal{W}$). For each group $g_i$ we set $\beta_{g_i} = 2$ - ensuring that both clusters must be $\alpha$-represented by the group. We also create one group for each clause $K_i$ corresponding to its three conditions $g_{K_i} = \{x_{v_{i1}}, x_{v_{i2}}, x_{v_{i3}}\}$ that must be $\alpha$-represented in cluster $1$ (i.e. $(g_{K_i},1) \in \mathcal{W}, (g_{K_i},2) \notin \mathcal{W}$). For these groups we set $\beta_{K_i} = 1$. Finally, we set $\alpha = \frac{1}{2n}$ - this ensures that any assignment of a group's data point to a cluster will satisfy the $\alpha$-representation constraint (as there are $2n$ data points and thus at most $2n$ data points in a cluster). We also add a cardinality lower bound on both clusters of 1. Clearly the above scheme can be set-up in polynomial time.

We now claim that a feasible solution to the aforementioned fair assignment problem corresponds to a solution to the original 3-SAT instance. We start by taking the variable settings by looking at $C_1$. We start by claiming that for each variable exactly one of $x_{v_i}, x_{\bar{v}_i}$ are included in $C_1$. Suppose this weren't true, either both variables were included in $C_1$ or $C_2$ - however whichever cluster has neither of the variables would not be $\alpha$-represented by group $g_i$ contradicting the constraints. Since $C_1$ contains either $x_{v_i}$ or $x_{\bar{v}_i}$ we set $v_i = T$ if $x_{v_i}$ is included and $v_i = F$ otherwise. We now claim that such a setting of the variables satisfies all the clauses. Assume it did not, then there exists a clause $K_i$ such that none of $x_{v_{i1}}, x_{v_{i2}}, x_{v_{i3}}$ are included in $C_1$. However, this violates the $\alpha$-representation constraint for $g_{K_i}$ providing a contradiction to the feasibility of the fair assignment problem.
\end{proof}

\section{Proof for Theorem \ref{thm:finite_convergence}} \label{sec:proof_convergence}
\begin{proof}
We start by showing that Algorithm \ref{alg:minrepfairlloyd} returns a local optimum. Suppose this were not the case, then there must be a local move that could improve the objective - specifically either a perturbation of a cluster center, or changing a single data point's cluster assignment. Consider the first case - a perturbation to a cluster center improves the objective. It is easy to verify that for a given cluster the optimal center is the mean of every data point in the cluster by looking at first and second order optimality conditions. Consider the derivative of the loss function with respect to a cluster center $c_k$:

$$
\frac{d}{dc_k} \sum_{x^i \in C_k} \|x^i - c_k\|^2_2 = \sum_{x^i \in C_k} 2(c_k - x^i)
$$

By setting the derivative of the loss to $0$, we see that $c_k = \frac{1}{n}  \sum_{x^i \in C_k} x^i$ is a critical point. Furthermore, the loss function is convex confirming that this point is in fact a minimizer. Thus a local perturbation of any cluster center cannot result in a decrease to the objective.  However, changing any single data point's cluster assignment will also not decrease the objective since we solve the fair assignment problem to optimality via integer programming. Thus the solution must be a local optimum.

It remains to show that the algorithm will converge in a finite amount of time. Similar to the proof for Lloyd's algorithm we leverage the fact that there exists a finite number of partitions of the data points. By construction at each iteration of the algorithm we decrease the objective value, and thus can never cycle through any partition multiple times as for a given partition of the data set we use the optimal cluster centers (as proven above). Thus the algorithm in the worst case can visit each partition once and thus must terminate in finite time.

\end{proof}

\section{Additional results for initialization schemes} \label{sec:exp_inits}

\begin{figure}[!htb]
    \centering
    \begin{subfigure}
      \centering
      \includegraphics[width=\textwidth]{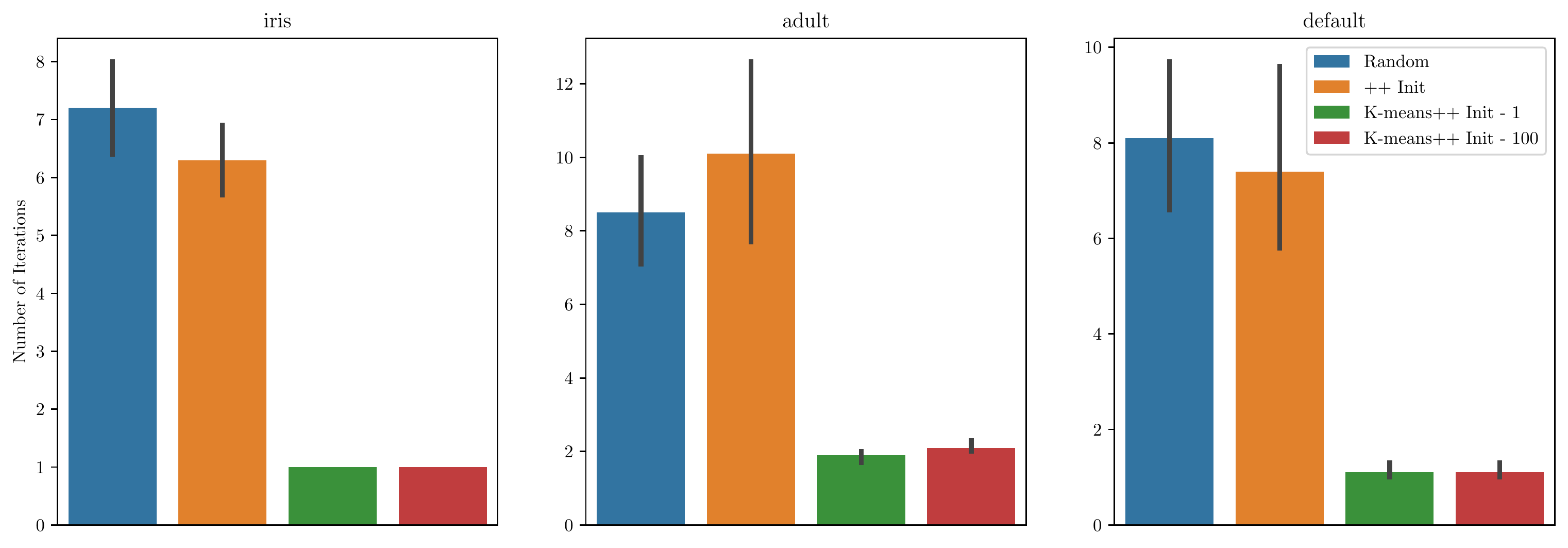}
    \end{subfigure}
  \caption {\label{fig:init_reps} Impact of cluster center initialization on the number of iterations to convergence for MiniReL. Results averaged over 10 random seeds. Bars indicate standard error.
  }
\end{figure}

\begin{figure}[!htb]
    \centering
    \begin{subfigure}
      \centering
      \includegraphics[width=\textwidth]{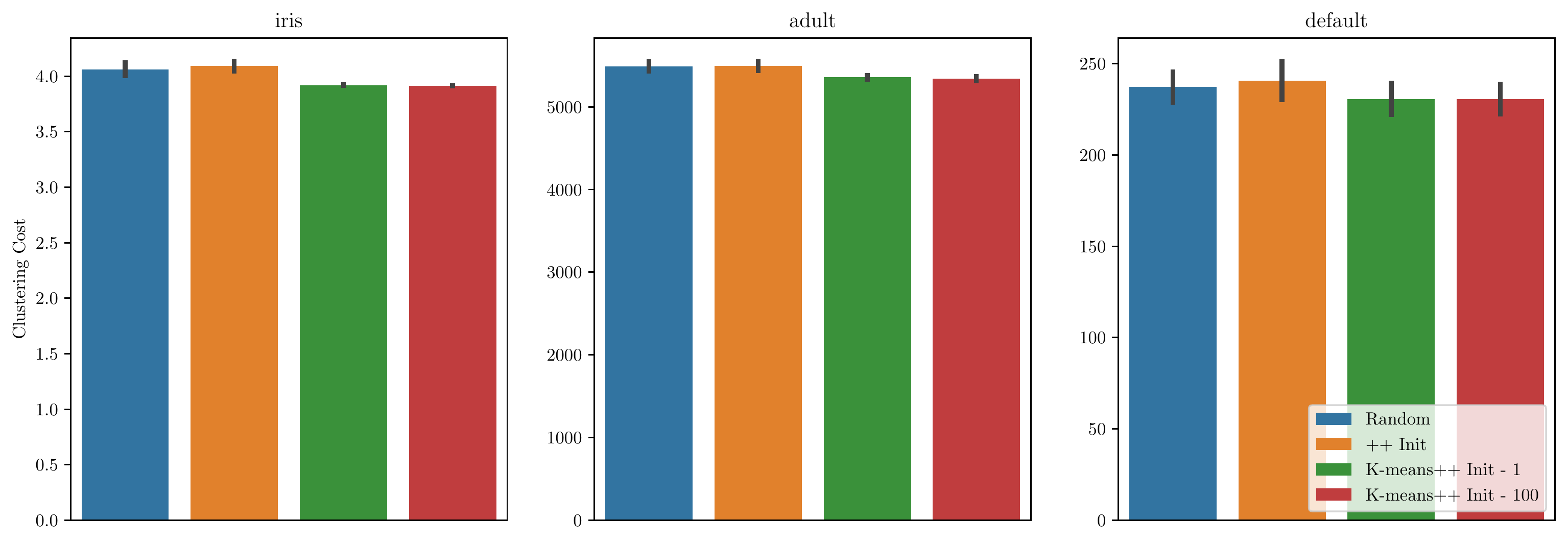}
    \end{subfigure}
  \caption {\label{fig:init_cost} Impact of cluster center initialization on the clustering cost of the solution output by MiniReL. Results averaged over 10 random seeds. Bars indicate standard error.
  }
\end{figure}

Figure \ref{fig:init_reps} shows the impact of initialization schemes on the number of repetitions needed for MiniReL to converge. Once again, warm-starting with $k$-means dramatically reduces the number of iterations needed compared to both random sampling and the $k$-means++ initialization. The number of random initializations (i.e. 1 vs. 100 initializations) for $k$-means has no impact on the number of iterations, showing that the increase in computation time is a consequence of the added overhead to compute the initial cluster centers, not necessarily the quality of the centers found. Figure \ref{fig:init_cost} presents the impact of initialization schemes on the quality of the final solution returned by MiniReL. Warm-starting with $k$-means leads to a small improvement in the quality of solution found (i.e. lower $k$-means cost), again signalling the computational benefits of warm-starting.  

\section{Additional results for pre-fixing} \label{sec:exp_prefix}
\begin{figure}[!htb]
    \centering
    \begin{subfigure}
      \centering
      \includegraphics[width=\textwidth]{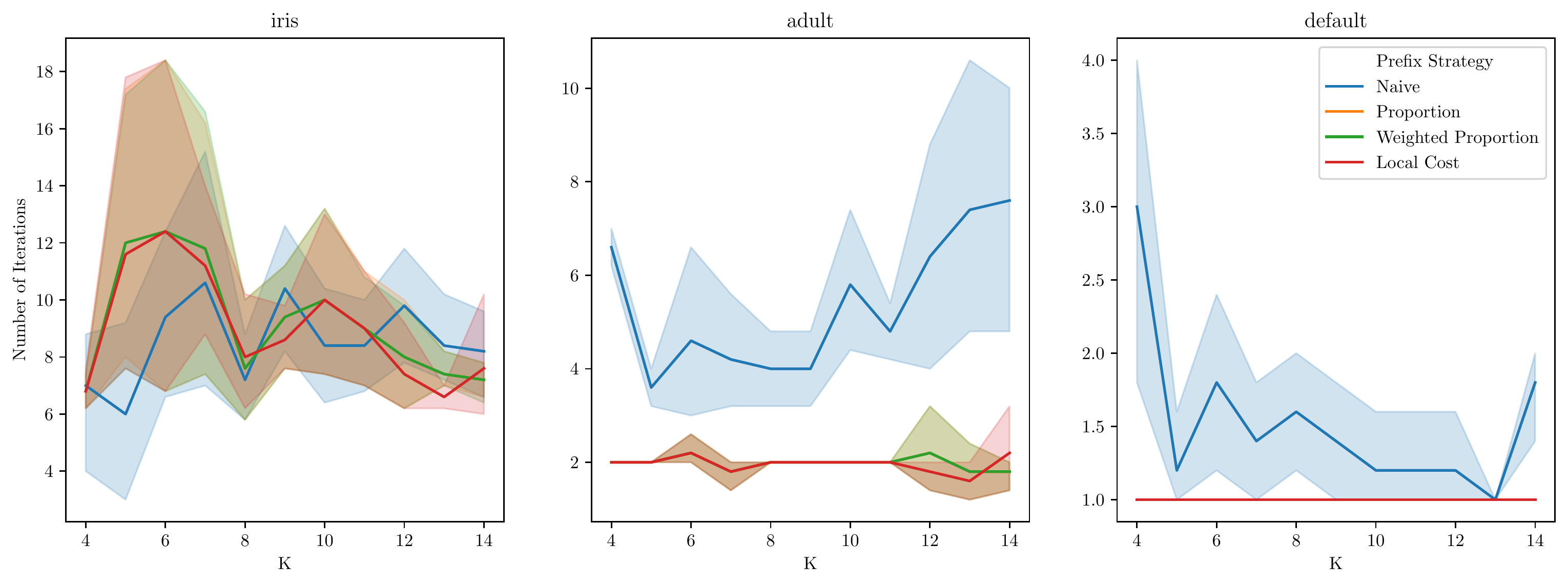}
    \end{subfigure}
  \caption {\label{fig:prefix_iter} Impact of pre-fixing strategy on the on the number of iterations to convergence for MiniReL. Results averaged over 10 random seeds. Bars indicate standard error.
  }
\end{figure}

\begin{figure}[!htb]
    \centering
    \begin{subfigure}
      \centering
      \includegraphics[width=\textwidth]{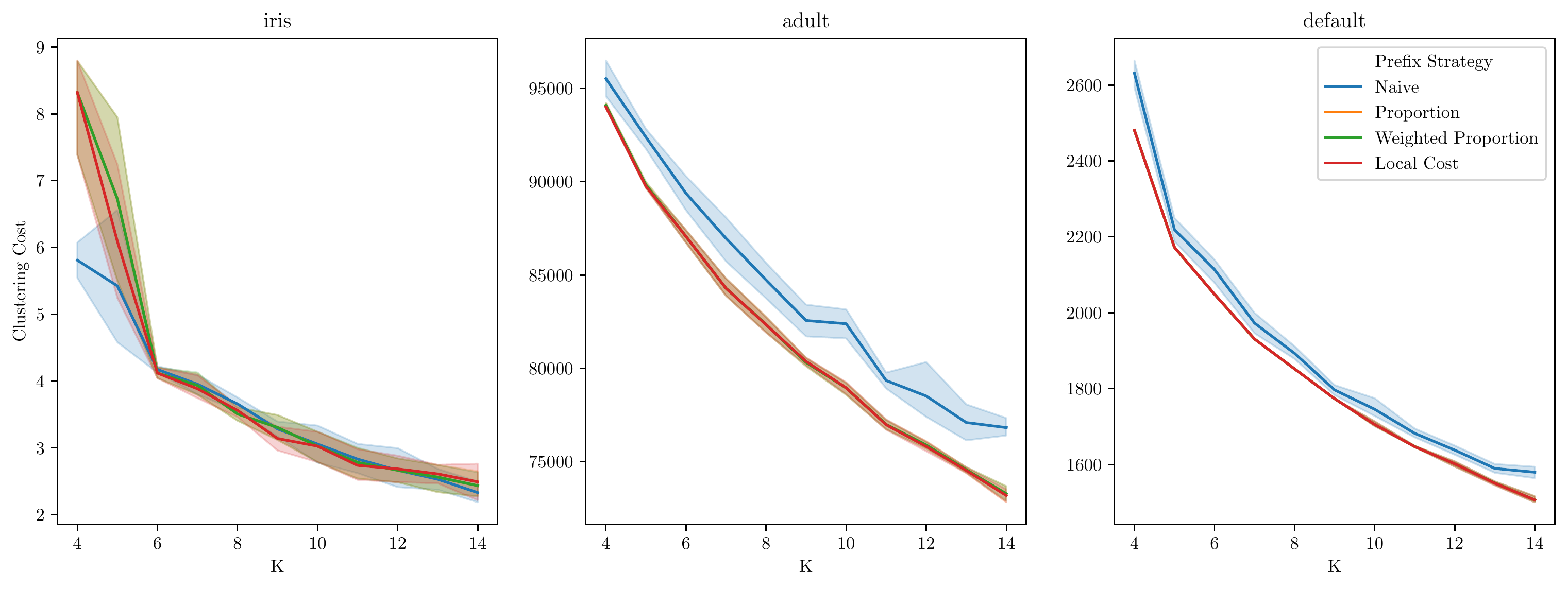}
    \end{subfigure}
  \caption {\label{fig:prefix_cost} Impact of pre-fix strategy on the clustering cost of the solution output by MiniReL. Results averaged over 10 random seeds. Bars indicate standard error.
  }
\end{figure}

\begin{figure}[!htb]
    \centering
    \begin{subfigure}
      \centering
      \includegraphics[width=\textwidth]{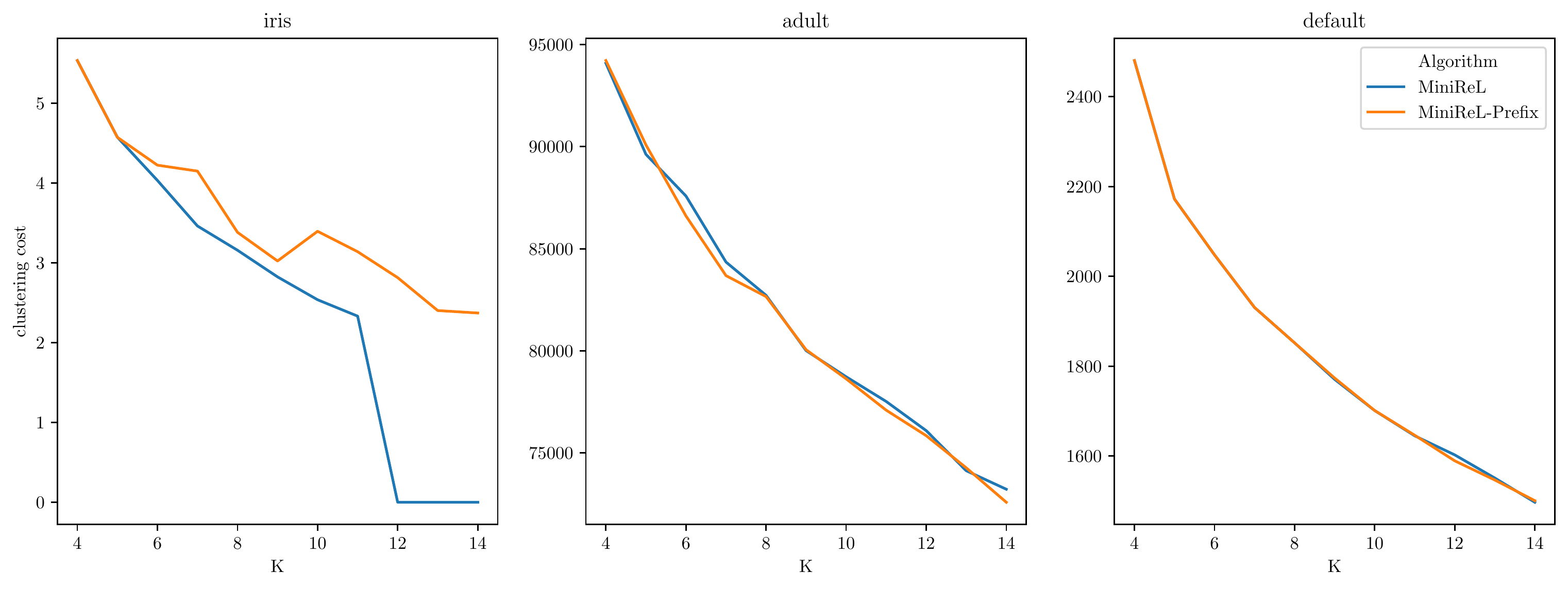}
    \end{subfigure}
  \caption {\label{fig:prefix_objective} Impact of pre-fixing on the clustering cost of the solution output by MiniReL.
  }
\end{figure}

Figure \ref{fig:prefix_iter} shows the impact of pre-fixing strategy on the number of iterations needed for MiniReL to converge. For iris, pre-fixing has practically no impact on on the number of iterations, however for larger datasets like adult and default using the pre-fix IP model with any objective leads to substantially fewer iterations. The same holds for cluster cost as shown in Figure \ref{fig:prefix_cost} where the IP model leads to solutions with slightly better clustering cost in both adult and default. Both results show that the choice of objective function has relatively little impact on the performance of pre-fixing, but outperform random assignment. Figure \ref{fig:prefix_objective} shows the impact of pre-fixing on the clustering cost of the final solution returned by MiniReL. Here we can see, with the exception of iris, pre-fixing has practically no impact on the clustering cost showing that the time savings come at no additional cost.


\end{document}